\documentclass{article}

\usepackage{PRIMEarxiv}

\usepackage[utf8]{inputenc} 
\usepackage[T1]{fontenc}    
\usepackage{hyperref}       
\usepackage{url}            
\usepackage{booktabs}       
\usepackage{amsfonts}       
\usepackage{amssymb}        
\usepackage{nicefrac}       
\usepackage{microtype}      
\usepackage{lipsum}
\usepackage{fancyhdr}       
\usepackage{graphicx}       
\usepackage{amsmath}
\usepackage{xspace}         
\graphicspath{{media/}}     

\makeatletter
\DeclareRobustCommand\onedot{\futurelet\@let@token\@onedot}
\def\@onedot{\ifx\@let@token.\else.\null\fi\xspace}

\def\eg{\emph{e.g}\onedot}

\makeatother

\title{SA-EMO: Structure-Aligned Encoder Mixture of Operators for Generalizable Full-waveform Inversion}

\author{
  Zhenyu Wang\textsuperscript{2}\thanks{These authors contributed equally.},
  Peiyuan Li\textsuperscript{1}\footnotemark[1],
  Yongxiang Shi\textsuperscript{1}\footnotemark[1],
  Ruoyu Wu\textsuperscript{3},
  Lei Zhang\textsuperscript{1}\thanks{Corresponding author.} \\
  \textsuperscript{1}School of Science, China University of Mining and Technology, Beijing \\
  \textsuperscript{2}School of Artificial Intelligence, China University of Mining and Technology, Beijing \\
  \textsuperscript{3}City University of Hong Kong (Dongguan)
}

\begin{document}
\maketitle

\begin{abstract}
Full-waveform inversion (FWI) can produce high-resolution subsurface models, yet it remains inherently ill-posed, highly nonlinear, and computationally intensive. Although recent deep-learning and numerical-acceleration methods have improved speed and scalability, they often rely on single-CNN architectures or single neural operators, which struggle to generalize in unknown or complex geological settings and are ineffective at distinguishing diverse geological types. To address these issues, we propose an Structure-Aligned Encoder–Mixture-of-Operators (SA-EMO) architecture for velocity-field inversion under unknown subsurface structures. First, a structure-aligned encoder maps high-dimensional seismic wavefields into a physically consistent latent space, thereby eliminating spatio-temporal mismatch between the waveform and velocity domains, recovering high-frequency components, and enhancing feature generalization. Then, an adaptive routing mechanism selects and fuses multiple neural-operator experts—including spectral, wavelet, multiscale, and local operators—to predict the velocity model. We systematically evaluate our approach on the OpenFWI benchmark and the Marmousi2 dataset. Results show SA-EMO significantly outperforms traditional CNN or single-operator methods, achieving an average MAE reduction of approximately 58.443\% and an improvement in boundary resolution of about 10.308\%. Ablation studies further reveal that the structure-alignment encoder, the expert-fusion mechanism, and the routing module each contribute markedly to the performance gains. This work introduces a new paradigm for efficient, scalable, and physically interpretable full-waveform inversion.
\end{abstract}


\section{Introduction}
\begin{figure}[t]
    \vspace{-6pt}
    \centering
    \includegraphics[width=0.46\textwidth]{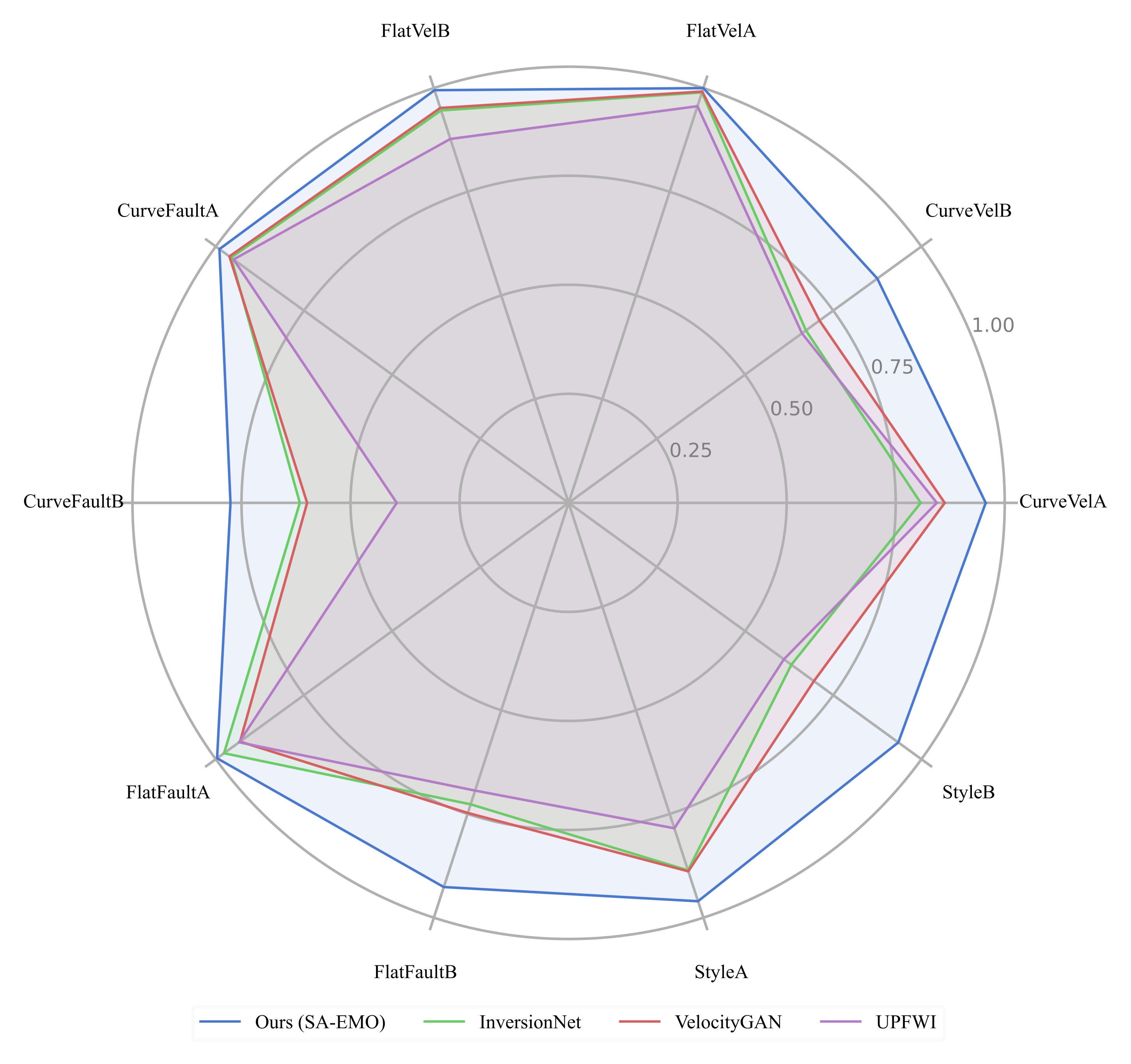}
    \vspace{-6pt}
    \caption{Cross-domain SSIM comparison of data-driven inversion frameworks.}
    \label{fig:rader}
    \vspace{-8pt}
\end{figure}
Full waveform inversion (FWI) aims to reconstruct high-fidelity subsurface velocity structures from multi-source seismic wavefields. Despite remarkable speedups achieved by deep learning, existing models still exhibit inconsistent cross-type generalization and degraded high-frequency reconstruction under complex geological conditions. We trace this limitation to an overlooked factor: the \emph{temporal–spatial and structural misalignment} between the waveform domain (time $\times$ receivers) and the velocity domain (2D structure). This mismatch causes uneven spectral energy distribution, degrading both learning stability and boundary fidelity.

Spectral analysis reveals that naive geometric interpolation (\eg, resizing wavefields onto the velocity grid) superficially aligns shapes but distorts frequency components and erases fine details, leading to over-smoothing and poor boundary recovery. Consequently, even powerful single-operator architectures such as FNO or WNO struggle to generalize across diverse geological structures.

To address this challenge, we present the Structure-Aligned Encoder–Mixture-of-Operators (SA-EMO) framework. The design consists of three tightly coupled components. 
First, a \emph{structure-aligned encoder} learns a physically consistent projection from raw wavefields to a $70{\times}70$ latent space that aligns with the target velocity structure. 
Second, four \emph{complementary neural operator experts}—FNO, WNO, MNO, and LNO—jointly capture global, multiscale, hierarchical, and local physical priors within this latent space. 
Third, an \emph{adaptive routing mechanism} fuses experts via type-weighted and group-weighted aggregation with \emph{strong–weak activation}, dynamically balancing dominant and auxiliary experts to adapt to different geological regimes. 
The entire system is trained with a hybrid spatial–spectral loss for unified optimization.

\textbf{Empirically}, the encoder primarily improves per-type inversion accuracy, whereas the routing mechanism enhances cross-type generalization. On OpenFWI~\cite{openfwi}, coupling FNO with our encoder improves MAE/RMSE/SSIM from $0.104/0.147/0.903$ to $0.085/0.125/0.925$ on CurveVelB. Strong–weak routing further enhances robustness across all ten sub-datasets and preserves structural continuity on the Marmousi2~\cite{marmousi} dataset, demonstrating strong out-of-distribution (OOD) generalization.

\paragraph{Contributions.}
\begin{itemize}
    \item We propose a \emph{structure-aligned encoder} that explicitly aligns waveform and velocity domains, achieving balanced spectral energy and improved single-operator accuracy.
    \item We introduce an \emph{operator-level mixture-of-experts} that integrates FNO, WNO, MNO, and LNO to jointly capture global, multiscale, and local physical priors in a unified latent space.
    \item We design a \emph{strong–weak activation routing mechanism} within a hierarchical type- and group-weighted framework, improving cross-type generalization and interpretability without explicit type labels.
    \item We conduct extensive experiments on ten OpenFWI sub-datasets and the OOD Marmousi2 dataset, showing that the encoder improves accuracy while routing enhances generalization.
\end{itemize}

\section{Related Work}
\label{sec:relatedwork}

In this section, we review prior work related to our method, covering physics-based full waveform inversion, deep learning-based inversion, neural operator learning and its applications to FWI, encoder–neural-operator architectures, and mixture-of-experts models. We highlight their limitations and explain how our framework builds upon and extends these directions.

\subsection{Physics-based Full Waveform Inversion}

Classical full waveform inversion (FWI) iteratively minimizes the data misfit between observed and simulated wavefields under a physical wave equation~\cite{Virieux2009}.  
Although these approaches provide physically interpretable reconstructions, they suffer from non-convexity, sensitivity to initial models, and high computational cost~\cite{Geng2018,Song2023,DesigningFWI2023}.  
Recent reviews~\cite{Operto2022,UnlockingFWI2023,ElasticFWI2025} summarize extensions beyond the Born approximation and emphasize the need for scalable, data-efficient solutions.  
However, traditional FWI still struggles in complex geological settings due to the ill-posed nature of the inversion problem and strong nonlinearity of wave propagation.

\subsection{Deep Learning for Seismic Inversion}

With the rise of data-driven modeling, deep learning has been introduced to approximate the nonlinear FWI mapping directly from seismic data to velocity fields~\cite{araya2018deep}.  
CNN-based architectures such as InversionNet~\cite{wu2019inversionnet}, VelocityGAN~\cite{zhang2019velocitygan}, and UPFWI~\cite{yang2023fwigan} achieve remarkable inference speed and robustness.  
However, they often exhibit poor generalization to unseen geological patterns~\cite{zhang2020data}.  
Hybrid frameworks that embed physical constraints or forward modeling within neural networks~\cite{jin2021unsupervised,chen2020seismic,jin2024empirical,schuster2024review} improve stability and interpretability but remain limited in scalability and adaptability across multiple geologies.

\subsection{Neural Operators and Applications to Full Waveform Inversion}

Neural operators (NOs) generalize function mappings between infinite-dimensional spaces, enabling resolution-independent PDE solvers~\cite{kovachki2023neural}.  
Representative families include the Fourier Neural Operator (FNO)~\cite{li2023fourier}, the Wavelet Neural Operator (WNO)~\cite{tripura2023wavelet}, the Multiscale Neural Operator (MNO)~\cite{lutjens2022multiscale}, and the Local Neural Operator (LNO)~\cite{li2024local}.  
These models achieve superior efficiency and generalization in many physical domains, inspiring their adoption for seismic inversion and wavefield modeling.  

Recent works have demonstrated that NOs can act as differentiable surrogates for forward solvers, accelerating physics-based FWI by orders of magnitude~\cite{zou2025ambient,yang2023rapid}.  
Others employ FNO or DeepONet variants to directly learn the inverse mapping from seismic wavefields to velocity structures~\cite{zhu2023fourier,gelboim2022encoder,taufik2025latent}.  
Although these studies confirm the potential of operator learning in FWI, they typically rely on single-operator architectures, which struggle with complex, heterogeneous subsurfaces where frequency content and structural scales vary dramatically.  
This motivates exploring modular operator ensembles and domain-aware latent conditioning, as proposed in our framework.

\subsection{Encoder\mbox{-}Neural Operator Architectures}
Encoder–decoder paradigms have been widely integrated with neural operators to bridge high-dimensional data and function-space mappings.  
U-shaped operators such as U-NO and U-FNO~\cite{rahman2022uno,wen2022ufno} employ convolutional encoders for multiscale spatial extraction, while Variational Autoencoding Neural Operators (VANO)~\cite{seidman2023variational} and Differentiable Autoencoding Neural Operators (DA-NO)~\cite{viknesh2025differentiable} embed operator learning in a latent space for uncertainty modeling.  
Mesh-Informed Neural Operators (MINO)~\cite{shi2025mesh} use Transformer encoders to encode geometric and sampling structures, and U-WNO~\cite{lei2025uwno} combines wavelet representations with U-Net encoders for multiscale feature preservation.  
Despite these advances, most existing encoders focus either on spatial hierarchy or on frequency balancing, lacking explicit temporal–spatial alignment between waveform and velocity domains.  
This limitation is critical in seismic FWI, where waveforms $\mathbf{S}\!\in\!\mathbb{R}^{5\times1000\times70}$ and velocity maps $\mathbf{V}\!\in\!\mathbb{R}^{70\times70}$ are inherently misaligned.  

Our work addresses this gap by introducing a structure-aligned encoder that explicitly harmonizes energy and geometry before operator learning.  
Leveraging the self-distillation and global self-attention mechanisms of DINOv3~\cite{simeoni2025dinov3}, the encoder redistributes waveform energy across temporal–spatial tokens to achieve spectral balance and structural consistency, forming a latent space $\mathbf{Z}$ that is physically coherent and directly compatible with operator inference.

\subsection{Mixture-of-Experts and Modular Adaptation}

Mixture-of-Experts (MoE) architectures~\cite{jordan1994hierarchical} combine specialized subnetworks under a routing or gating mechanism, improving capacity and adaptability.  
This principle has powered large-scale sparse models such as Switch Transformers~\cite{fedus2022switch}, GShard~\cite{lepikhin2020gshard}, and GLaM~\cite{du2022glam}, and has recently been extended to vision~\cite{riquelme2021scaling} and scientific computing~\cite{bischof2022mixture,hu2023augmented,sharma2024ensemble}.  
Within PDE learning, modular experts enable decomposition over physical domains or basis functions, improving generalization and interpretability.  
Inspired by these advances, our work introduces a neural operator-level MoE framework where each expert specializes in distinct physical priors (spectral, wavelet, multiscale, or local).  
A latent-domain router adaptively fuses expert predictions based on the encoded structure, yielding an interpretable and adaptive operator ensemble.

\subsection{Summary and Positioning of Our Work}
In summary, previous data-driven FWI frameworks either employ monolithic CNNs or single-operator architectures, limiting their generalization to unseen geological regimes.  
We bridge this gap by unifying encoder–latent alignment, multi-operator specialization, and adaptive expert routing into a single Encoder–Mixture-of-Operators (SA-EMO) architecture.  
This design merges physical interpretability with data-driven scalability, achieving robust and structure-aware inversion across diverse geological families in the OpenFWI~\cite{openfwi} benchmark.

\section{Methodology}
\label{sec:method}

\begin{figure*}[t]
    \centering
    \includegraphics[width=\linewidth]{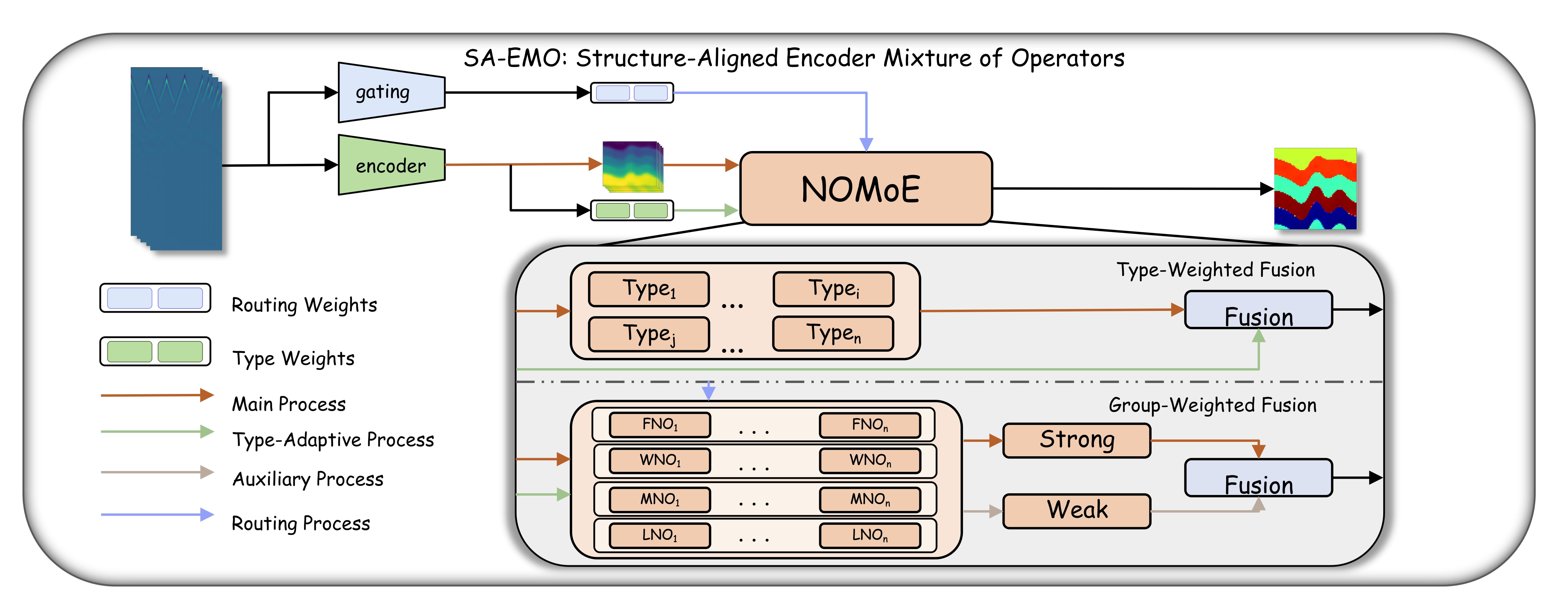}
    \caption{\textbf{Overview of the proposed Encoder–Mixture-of-Operators (SA-EMO) framework.} 
    SA-EMO integrates four synergistic components: 
    (1) a structure-aligned encoder that maps the seismic wavefield to a physically consistent latent domain; 
    (2) a set of complementary neural operator experts capturing global-to-local geological features; 
    (3) an adaptive routing mechanism for structure-aware expert fusion; and 
    (4) a hybrid spatial–spectral–type-aware loss for unified supervision.}
    \label{fig:framework}
\end{figure*}

As illustrated in Fig.~\ref{fig:framework}, our proposed Encoder–Mixture-of-Operators (SA-EMO) framework comprises four core modules: a structure-aligned encoder, complementary neural operator experts, an adaptive routing mechanism, and a hybrid spatial–spectral loss. 
Each component addresses limitations in prior FWI and neural operator methods~\cite{wu2019inversionnet,zhang2019velocitygan,li2023fourier,tripura2023wavelet,kovachki2023neural} 
and collectively enables physics-consistent and generalizable full-waveform inversion.

\subsection{Structure-Aligned Encoder}
\paragraph{Motivation.}
Traditional neural operators such as the Fourier Neural Operator (FNO)~\cite{li2023fourier} 
and the Wavelet Neural Operator (WNO)~\cite{tripura2023wavelet} 
directly process seismic wavefields 
$\mathbf{S}\in\mathbb{R}^{5\times1000\times70}$ 
to reconstruct velocity maps 
$\mathbf{V}\in\mathbb{R}^{70\times70}$.
However, the temporal–spatial misalignment between waveform and velocity domains leads to blurred predictions and poor generalization. 
Spectral analysis reveals that this mismatch induces an uneven frequency energy distribution, making operator learning unstable. 
Simply interpolating a waveform $\mathbf{S}\in\mathbb{R}^{1000\times350}$ onto a $70\times70$ grid provides only superficial alignment but destroys high-frequency energy, resulting in over-smoothing and degraded boundary resolution, consistent with prior findings in physics-guided inversion~\cite{gelboim2022encoder,taufik2025latent}. 

\paragraph{Encoder Architecture.}
To overcome these issues, we need an encoder that can jointly balance spectral energy and establish spatial alignment between seismic wavefields and velocity maps.  
This requires two essential capabilities: 
(1) capturing long-range temporal dependencies while preserving waveform continuity, and 
(2) learning global spatial relationships to project the wavefield into a topology-consistent 2D representation.  
The self-distillation Vision Transformer DINOv3~\cite{simeoni2025dinov3} naturally satisfies both requirements through its global self-attention and multi-scale token interaction mechanisms.  
Its attention layers dynamically redistribute information across time and space, while the [CLS] token encodes global structure without explicit interpolation.  

Therefore, we employ DINOv3 as the core of our structure-aligned encoder $E_\theta$, adapting it to the seismic setting as follows:
\begin{equation}
\mathbf{Z} = E_\theta(\mathbf{S}), \quad 
\mathbf{S}\in\mathbb{R}^{N_s\times T\times R}, \ 
\mathbf{Z}\in\mathbb{R}^{C\times H\times W},
\end{equation}
where $N_s=5$, $T=1000$, and $R=70$ denote the number of sources, temporal length, and receivers, while $H=W=70$ defines the latent spatial resolution.  
In practice, the temporal–receiver signals are tokenized into 2D patch embeddings and globally aggregated through attention, allowing each token to integrate both temporal propagation and spatial correlation.  
This produces a latent representation $\mathbf{Z}$ that exhibits \emph{spectral balance}, \emph{geometric consistency}, and a clear structural correspondence with the target velocity field.  
Hence, rather than serving as a generic vision backbone, DINOv3 functions here as an information–energy harmonizer that bridges the temporal–spatial gap and provides a physics-coherent latent space for operator learning.

\paragraph{Spectral Alignment.}
We compute the averaged Fourier spectra of the input and latent features, $\mathcal{F}(\mathbf{S})$ and $\mathcal{F}(\mathbf{Z})$. 
Results indicate that $\mathbf{Z}$ effectively restores high-frequency components near the Nyquist limit and exhibits a balanced spectral energy distribution, confirming frequency-consistent structural alignment. 
The latent representation $\mathbf{Z}$ thus provides a unified, physics-coherent input space for operator learning.

\paragraph{Physics-Guided Type Awareness.}
Beyond geometric alignment, the encoder learns physics-guided type awareness. 
We introduce an auxiliary prediction head $h_\phi$ that infers a geological type-weight vector from $\mathbf{Z}$:
\begin{equation}
\mathbf{w}_{\text{type}} = h_\phi(\mathbf{Z}), \quad 
\mathbf{w}_{\text{type}} \in [0,1]^K, \quad 
\sum_{k=1}^{K} w_{\text{type},k} = 1,
\end{equation}
where $K$ denotes the number of geological velocity families in the dataset (\eg, CurveVel, FlatVel, Fault, Style~\cite{jin2024empirical}). 
Each component $w_{\text{type},k}$ reflects the encoder’s confidence that $\mathbf{S}$ originates from type $k$. 
This mechanism enables unsupervised discovery of geological priors, guiding expert routing without explicit type labels~\cite{bischof2022mixture,hu2023augmented}.

\subsection{Complementary Neural Operator Experts}
\label{sec:no}
Each neural operator expert $\mathcal{O}_i$ is designed to specialize in a distinct physical prior of seismic wave propagation, forming a complementary operator ensemble:
\begin{equation}
\hat{\mathbf{V}}_i = \mathcal{O}_i(\mathbf{Z}), \qquad 
\mathbf{Z}\in\mathbb{R}^{C\times H\times W}, \ 
\hat{\mathbf{V}}_i\in\mathbb{R}^{1\times H\times W}.
\end{equation}

\paragraph{Fourier Neural Operator (FNO).}Following~\cite{li2023fourier,kovachki2023neural}, FNO performs spectral convolutions in the Fourier domain:
\begin{equation}
\mathcal{O}_F(\mathbf{Z}) = \mathcal{F}^{-1}\!\left(W_F \odot \mathcal{F}(\mathbf{Z})\right),
\end{equation}
where $\mathcal{F}$ and $\mathcal{F}^{-1}$ denote the Fourier and inverse transforms, and $\odot$ indicates element-wise multiplication.  
FNO captures long-range dependencies and models smooth stratified layers via global spectral interactions.

\paragraph{Wavelet Neural Operator (WNO).}Inspired by~\cite{tripura2023wavelet}, WNO decomposes $\mathbf{Z}$ into multiple frequency scales using discrete wavelet transforms:
\begin{equation}
\mathcal{O}_W(\mathbf{Z}) = \mathcal{W}^{-1}\!\big(W_W \odot(\mathcal{W}(\mathbf{Z}))\big),
\end{equation}
where $\mathcal{W}$ and $\mathcal{W}^{-1}$ are wavelet transforms and $W_W$ denotes learnable scale weights.  
This enables spatially localized, multi-frequency representation, preserving anisotropic boundaries and geological discontinuities.

\paragraph{Multiscale Neural Operator (MNO).}The MNO~\cite{lutjens2022multiscale} expands receptive fields via hierarchical convolution kernels:
\begin{equation}
\mathcal{O}_M(\mathbf{Z}) = \sum_{s=1}^{S} \phi_s(K_s * \mathbf{Z}),
\end{equation}
where $K_s$ denotes scale-specific kernels and $\phi_s$ aggregates multi-resolution responses.  
This allows robust modeling of complex geological structures with spatially varying scales.

\paragraph{Local Neural Operator (LNO).}Following localized kernel formulations~\cite{li2024local}, LNO introduces position-dependent filters:
\begin{equation}
\mathcal{O}_L(\mathbf{Z})(x) = \int_{\Omega_x} K(x,x')\,\mathbf{Z}(x')\,dx' + \psi(\nabla\mathbf{Z}(x)),
\end{equation}
where $K(x,x')$ is constrained to a neighborhood $\Omega_x$ and $\psi(\cdot)$ captures local gradients.  
LNO enhances spatial adaptivity and reconstructs high-contrast discontinuities such as faults and channels.

\paragraph{}Collectively, $\{\mathcal{O}_F, \mathcal{O}_W, \mathcal{O}_M, \mathcal{O}_L\}$ form a physics-aware operator ensemble:  
FNO models global smoothness, WNO extracts multi-scale edges, MNO captures hierarchical structure, and LNO restores fine-grained discontinuities.  
All experts are pretrained on diverse velocity families and jointly fine-tuned within SA-EMO, ensuring spectral diversity and structural robustness.

\subsection{Adaptive Routing Mechanism}
\label{sec:routing}

\paragraph{Motivation.}While individual experts excel in distinct regimes, their relative importance varies with geological context. 
To achieve adaptive expert collaboration without explicit type labels, 
we propose a dual-stage adaptive routing mechanism that integrates type-weighted routing and group-weighted fusion, inspired by mixture-of-experts (MoE) frameworks~\cite{jordan1994hierarchical,shazeer2017outrageously,fedus2022switch,riquelme2021scaling,shi2024unchosen}.

\paragraph{Type-Weighted Routing.}For each geological type $t$, we first pretrain all experts $\{\mathcal{O}_i\}_{i=1}^4$ independently and evaluate their inversion performance.  
The best-performing expert $\mathcal{O}^*_t$ for each type is then selected and incorporated into the Mixture-of-Experts (MoE) framework with frozen parameters.  
During inference, the encoder produces a type-aware latent representation $\mathbf{Z}$, which is projected into a type-probability vector:
\begin{equation}
\boldsymbol{\gamma} = f_\theta(\mathbf{Z}), \quad
\boldsymbol{\gamma} \in [0,1]^T, \quad
\sum_{t=1}^{T}\gamma_t = 1,
\end{equation}
where $\gamma_t$ denotes the probability that the current sample belongs to type $t$.  
The final prediction is then obtained by weighting the outputs of the selected experts according to $\boldsymbol{\gamma}$:
\begin{equation}
\hat{\mathbf{V}} = \sum_{t=1}^{T} \gamma_t\, \mathcal{O}^*_t(\mathbf{S}),
\end{equation}
allowing the model to dynamically emphasize the expert most relevant to the inferred geological type.

\paragraph{Group-Weighted Fusion.}To retain all experts’ contributions, we organize them by inductive bias:
\begin{equation}
\mathcal{G}_1=\{\text{FNO}\},\;
\mathcal{G}_2=\{\text{WNO}\},\;
\mathcal{G}_3=\{\text{MNO}\},\;
\mathcal{G}_4=\{\text{LNO}\}.
\end{equation}
Each group produces a weighted intra-group prediction:
\begin{equation}
\hat{\mathbf{V}}_g = \sum_{t=1}^{T} \gamma_t\, \mathcal{O}_{g,t}(\mathbf{S}),
\end{equation}
resulting in four group-level outputs $\{\hat{\mathbf{V}}_g\}$ for hierarchical fusion.

\paragraph{Strong–Weak Activation Fusion.}Inspired by the Self-Contrast MoE (SCMoE)~\cite{shi2024unchosen}, 
we design a contrastive activation strategy to balance dominant and complementary groups. 
Given router scores $\boldsymbol{\beta}=R_\phi(\mathbf{Z})$, 
we identify top-$k$ strong activations ($k{=}2$) and remaining weak activations:
\[
\mathcal{S}=\text{TopK}(\boldsymbol{\beta},k), \quad
\mathcal{W}=\{1,2,3,4\}\setminus\mathcal{S}.
\]
After normalization within each set, the final fusion is:
\begin{equation}
\hat{\mathbf{V}} =
(1+\lambda)\sum_{g\in\mathcal{S}}\bar{\beta}_g^{\text{strong}}\hat{\mathbf{V}}_g
-
\lambda\sum_{g\in\mathcal{W}}\bar{\beta}_g^{\text{weak}}\hat{\mathbf{V}}_g,
\end{equation}
where $\lambda$ controls contrast intensity.
This strategy enables structure-aware expert activation, allowing both dominant and suppressed experts to contribute effectively under varying geological complexities.

\subsection{Hybrid Loss Function}
\label{sec:loss}

To ensure structural fidelity, spectral consistency, and physics-aware routing, 
SA-EMO is optimized with a hybrid objective:
\begin{equation}
\mathcal{L}_{\text{total}} =
\lambda_1 \mathcal{L}_{\text{struct}} +
\lambda_2 \mathcal{L}_{\text{Fourier}}.
\end{equation}

\paragraph{Spatial–Structural Supervision.}We apply combined L1–L2 reconstruction and first-order gradient losses:
\begin{align}
\mathcal{L}_{L1L2} &= \|\hat{\mathbf{V}}-\mathbf{V}\|_1 +
\|\hat{\mathbf{V}}-\mathbf{V}\|_2^2, \\
\mathcal{L}_{grad} &= \|\nabla_x(\hat{\mathbf{V}}-\mathbf{V})\|_1 +
\|\nabla_y(\hat{\mathbf{V}}-\mathbf{V})\|_1, \\
\mathcal{L}_{\text{struct}} &= \beta_1 \mathcal{L}_{L1L2} + \beta_2 \mathcal{L}_{grad},
\end{align}
which promote pixel-wise accuracy, edge preservation, and structural smoothness~\cite{wu2019inversionnet,zhang2020data}.

\paragraph{Spectral-Domain Supervision.}To maintain frequency-domain consistency, we employ the Fourier magnitude loss:
\begin{equation}
\mathcal{L}_{\text{Fourier}} =
\big\|\,|\mathcal{F}(\hat{\mathbf{V}})| - |\mathcal{F}(\mathbf{V})|\,\big\|_1,
\end{equation}
ensuring that reconstructed velocity models preserve both low- and high-frequency energy distributions~\cite{yang2023fwigan}.

\section{Experiments}
\label{sec:experiments}

We conduct extensive experiments on the OpenFWI benchmark to validate the effectiveness of our proposed Structure-Aligned Encoder –Mixture -of -Operators (SA-EMO) framework.  
Our study is organized progressively to answer three questions:  
(1) how the \emph{structure-aligned encoder} improves inversion accuracy,  
(2) how \emph{individual neural operators} differ in their inductive biases, and  
(3) how the \emph{routing mechanism} enhances cross-type generalization.  

\subsection{Experimental Setup}
\label{sec:exp-setup}
\paragraph{Datasets.}
We use all ten official 2D sub-datasets from OpenFWI~\cite{openfwi}, which cover diverse geological and fault conditions:
CurveVel\mbox{-}A/B, FlatVel\mbox{-}A/B, CurveFault\mbox{-}A/B, FlatFault\mbox{-}A/B, and Style\mbox{-}A/B.
Each sub-dataset provides pre-defined training and validation splits following the official configuration.
Specifically, the Vel family (FlatVel\mbox{-}A/B, CurveVel\mbox{-}A/B) contains 24k training and 6k validation pairs;
the Fault family (FlatFault\mbox{-}A/B, CurveFault\mbox{-}A/B) includes 48k training and 6k validation pairs;
and the Style family (Style\mbox{-}A/B) provides 60k training and 7k validation pairs.
Each sample consists of five seismic shot gathers
$\mathbf{S}\in\mathbb{R}^{5\times1000\times70}$
and a ground-truth velocity map
$\mathbf{V}\in\mathbb{R}^{1\times70\times70}$.
Since OpenFWI~\cite{openfwi} does not provide official normalization statistics, we compute the mean and standard deviation of $\mathbf{S}$ and $\mathbf{V}$ within each sub-dataset independently, and apply per-subset standardization before training to ensure consistent scale across input wavefields and velocity maps.
For each family, the “A’’ version represents an easier case with smoother geological structures,
while the “B’’ version introduces more complex and challenging patterns with higher-frequency components and stronger nonlinearity.
To further assess generalization, we conduct out-of-distribution (OOD) evaluation on the Marmousi2 dataset~\cite{marmousi}.

\paragraph{Implementation.}
All models are implemented in PyTorch~2.8. 
For the OpenFWI datasets, each seismic sample is reshaped by concatenating the shot dimension along the receiver axis, 
transforming $\mathbf{S}\in\mathbb{R}^{5\times1000\times70}$ into $\mathbf{S}'\in\mathbb{R}^{1\times1000\times350}$ 
to better capture temporal–spatial correlations before encoding. 
The encoder is trained using mixed precision (bf16), whereas all operator experts are trained in full precision (fp32). 
We employ the AdamW optimizer ($\text{lr}=2.6\times10^{-4}$, $\text{weight decay}=0.089$) 
with a WarmupCosineAnnealingWarmRestarts scheduler ($\gamma=0.29$). 
Training is conducted on dual RTX~4090 GPUs with a per-GPU batch size of 32 for 180~epochs. 
The hybrid loss function adopts weighting coefficients 
$\lambda_{\text{g1v}}=0.4395$, 
$\lambda_{\text{g2v}}=0.3534$, 
$\lambda_{\text{grad}}=0.15$, and 
$\lambda_{\text{Fourier}}=0.05$.
We first pre-train each expert operator individually following standard PyTorch protocols.
During the SA-EMO fine-tuning stage, the expert operators are frozen, while the encoder and routing network are jointly fine-tuned 
to adapt the latent representation and expert selection strategy for cross-type generalization.
Detailed hyperparameters for both single-expert pretraining and encoder pretraining are provided in the supplementary material.

\paragraph{Metrics.}
We report MAE (Mean Absolute Error), RMSE (Root Mean Square Error), and SSIM (Structural Similarity Index), following OpenFWI~\cite{openfwi} evaluation protocols.

\subsection{Effect of Structure-Aligned Encoder}
\label{sec:exp-encoder}

\paragraph{Rationale.}
We begin by analyzing the encoder, which forms the foundation for all subsequent operator and MoE experiments.
The encoder aligns the seismic waveform domain with the target velocity domain, enabling consistent spectral–spatial mapping for operator learning.

We evaluate the effectiveness of the proposed structure-aligned encoder on the challenging CurveVel\mbox{-}B dataset,
which contains highly curved geological layers and strong high-frequency reflections.
FNO serves as the fixed backbone, while four encoder configurations are compared:
None (original) directly inputs the raw waveform $\mathbf{S}\!\in\!\mathbb{R}^{1\times1000\times350}$ without resizing;
None (resize) interpolates the waveform to match the velocity grid ($70\times70$);
ViT\mbox{-}based and ConvNeXt\mbox{-}based encoders adopt the DINOv3 framework~\cite{simeoni2025dinov3} for structure\mbox{-}aligned representation learning.

\begin{figure}[t]
    \centering
    \includegraphics[width=\linewidth]{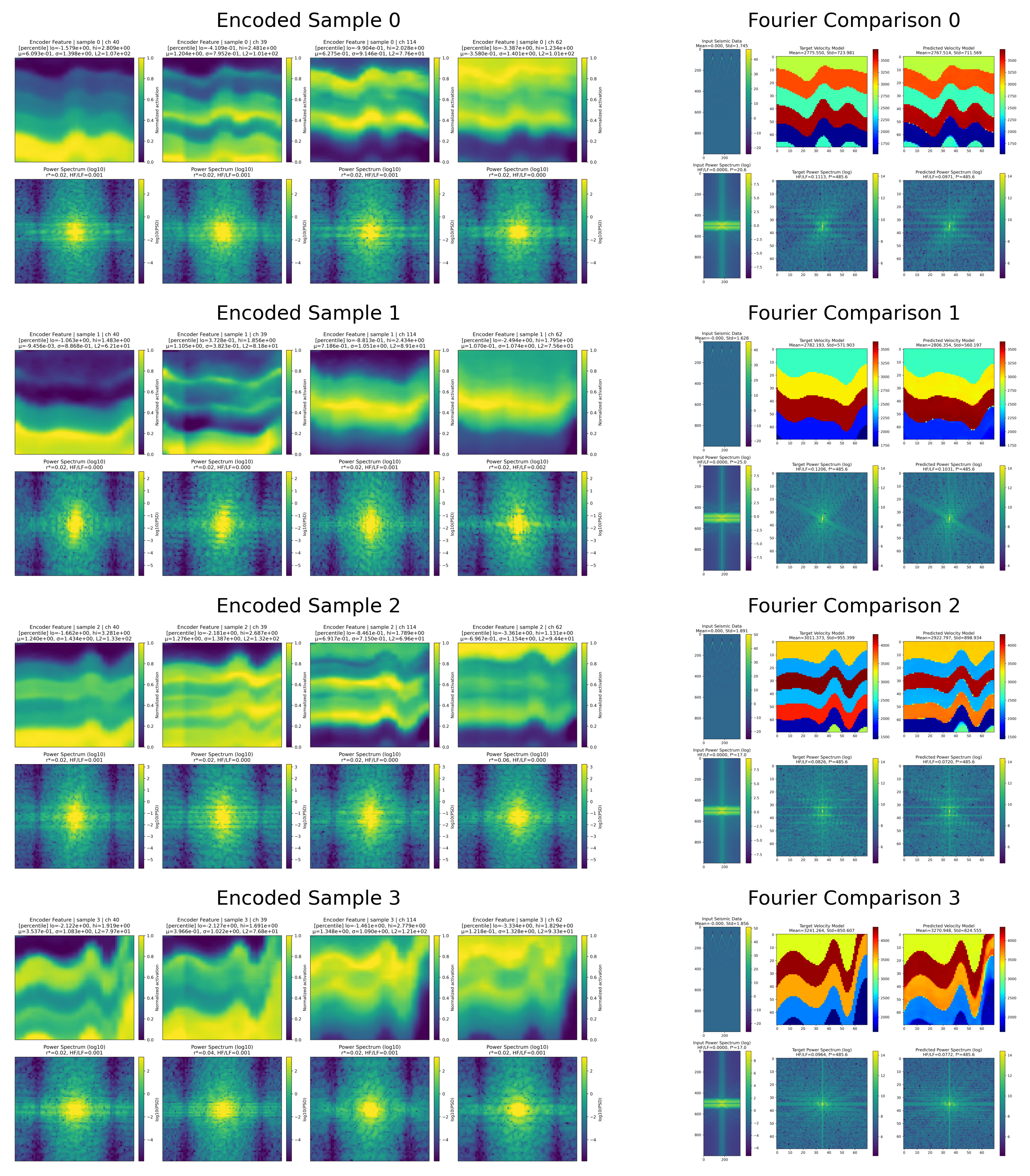}
    \caption{
    \textbf{Visualization of structure-aligned encoder and Fourier spectra on CurveVel-B.}
    Each row corresponds to one test sample.
    The first four columns show the encoder feature maps (channels 40, 39, 114, and 62) exhibiting layer-consistent activations;
    the next columns show their power spectra (log$_{10}$ scale), dominated by low frequencies ($r^{\ast}\!\approx\!0.02$) with HF/LF$<0.1$.
    The rightmost panels compare input, target, and predicted spectra, demonstrating perfect frequency alignment ($f_\text{pred}=f_\text{gt}=485.6$).
    }
    \label{fig:encoder_vis}
\end{figure}

\paragraph{Spectral and structural analysis.}
To further verify that the proposed encoder truly aligns the seismic and velocity domains,
we analyze its latent features and corresponding Fourier spectra across multiple test samples.
As shown in Fig.~\ref{fig:encoder_vis}, the structure-aligned encoder produces spatial activations that follow stratigraphic curvature
and exhibit smooth, layer-consistent patterns,
in contrast to the irregular textures obtained without encoding.
Quantitatively, the encoded channels exhibit extremely low high-frequency–to-low-frequency ratios (HF/LF $\approx 5{\times}10^{-2}$–$9{\times}10^{-2}$)
and dominant normalized frequencies $r^{\ast}\!\approx\!0.02$,
indicating that the encoder suppresses geometry-dependent noise and preserves physically meaningful low-frequency components
that correspond to subsurface layering.
Meanwhile, inter-channel correlations reveal clear complementarity:
spatial correlations fluctuate between $0.997$–$0.999$,
while spectral correlations remain almost perfect ($\approx 0.9999$–$1.0000$),
suggesting that each latent channel captures distinct spatial details yet shares a consistent global frequency structure.
This validates that the encoder establishes a stable spectral basis for subsequent operator learning.

\noindent \textbf{Definition of $r^{\ast}$.} 
For clarity, the normalized dominant frequency $r^{\ast}$ is computed as
$r^{\ast} = f_{\text{dom}} / f_{\text{max}}$,
where $f_{\text{dom}}$ denotes the radial distance of the spectral peak
and $f_{\text{max}}=\sqrt{(H/2)^2+(W/2)^2}$ is the theoretical Nyquist limit of the $H{\times}W$ grid.
Since the Fourier energy is scaled by the transform size,
the measured dominant frequency $f_{\text{dom}}\!\approx\!485.6$ (for $H{=}W{=}70$)
corresponds to a normalized value $r^{\ast}\!\approx\!0.02$,
indicating that most energy is concentrated within the lowest 2\% of the spectral range.

\begin{table}[t]
    \centering
    \caption{\textbf{Effect of structure-aligned encoder on CurveVelB (FNO backbone).}
    The ViT\mbox{-}based and ConvNeXt\mbox{-}based encoders are implemented using DINOv3~\cite{simeoni2025dinov3}
    and trained under identical settings.}
    \label{tab:encoder}
    \resizebox{\columnwidth}{!}{
    \begin{tabular}{lccc}
        \toprule
        Encoder Type & MAE $\downarrow$ & RMSE $\downarrow$ & SSIM $\uparrow$ \\
        \midrule
        None (original, $1000{\times}350$) & 0.118 & 0.161 & 0.891 \\
        None (resize to $70{\times}70$)     & 0.1863 & 0.3112 & 0.6151 \\
        ConvNeXt\mbox{-}based (DINOv3)      & 0.0637 & 0.1693 & 0.8634 \\
        ViT\mbox{-}based (DINOv3)           & \textbf{0.0598} & \textbf{0.1649} & \textbf{0.8637} \\
        \bottomrule
    \end{tabular}}
\end{table}

\paragraph{Fourier-domain validation.}
Fourier-domain comparison between input, target, and predicted velocity fields further confirms the alignment effect.
Across four CurveVel-B samples, the predicted spectra nearly coincide with the target spectra—
their dominant frequencies match exactly ($f_{\text{pred}}=f_{\text{gt}}=485.6$)
and spectral correlations reach above $0.9999$,
while direct waveform inputs differ by over an order of magnitude in spectral energy.
The encoder thus transforms the irregular waveform spectrum (dominated by temporal harmonics $\approx20$–$25$\,Hz)
into a velocity-like distribution centered on physically relevant spatial frequencies.

\begin{table}[t]
    \centering
    \caption{\textbf{Quantitative spectral metrics of encoder features across samples.}
    Low HF/LF ratios and near-unity spectral correlations confirm the encoder’s spectral–spatial consistency.}
    \label{tab:spectral_metrics}
    \resizebox{\columnwidth}{!}{
    \begin{tabular}{cccccc}
        \toprule
        Sample & HF/LF $\downarrow$ & $r^{\ast}$ & Spatial Corr & Spectral Corr & $E_{\text{pred}}/E_{\text{gt}}$ \\
        \midrule
        0 & 0.094 & 0.02 & 0.999998 & 0.999998 & 0.5049 \\
        1 & 0.094 & 0.02 & 0.999997 & 0.999997 & 0.5007 \\
        2 & 0.077 & 0.02 & 0.999997 & 0.999997 & 0.4971 \\
        3 & 0.077 & 0.02 & 0.999996 & 0.999996 & 0.5029 \\
        \bottomrule
    \end{tabular}}
\end{table}

\paragraph{Take-away.}
Direct interpolation provides limited improvement as it distorts spectral characteristics and loses spatial details.
In contrast, both DINOv3-based encoders substantially reduce MAE/RMSE while improving SSIM.
The ViT-based encoder achieves the best performance, confirming that
\emph{structure-aligned encoding effectively bridges the waveform–velocity domain gap.}

\subsection{Single-Expert Analysis}
\label{sec:exp-single-expert}

\begin{table*}[t]
\centering
\caption{\textbf{Performance of single operator experts on 10 OpenFWI~\cite{openfwi} sub-datasets (MAE$\downarrow$, RMSE$\downarrow$, SSIM$\uparrow$).}
Different operators exhibit distinct inductive biases.
Best per column in \textbf{bold}.}
\resizebox{\textwidth}{!}{
\begin{tabular}{lccccccccccc}
\toprule
Metric & CurveVelA & CurveVelB & FlatVelA & FlatVelB & CurveFaultA & CurveFaultB & FlatFaultA & FlatFaultB & StyleA & StyleB & Avg. \\
\midrule
\multicolumn{12}{l}{\textbf{MAE ↓}} \\
FNO~\cite{li2023fourier} & \textbf{0.0182} & \textbf{0.0534} & 0.0015 & 0.0073 & 0.0087 & 0.0835 & 0.0047 & 0.0349 & 0.0302 & \textbf{0.0338} & 0.0272 \\
WNO~\cite{tripura2023wavelet} & 0.2344 & 0.2540 & 0.2395 & 0.2388 & 0.2176 & 0.1945 & 0.2243 & 0.1607 & 0.0836 & 0.0809 & 0.2028 \\
MNO~\cite{lutjens2022multiscale} & 0.0198 & 0.0539 & \textbf{0.0013} & 0.0071 & \textbf{0.0078} & \textbf{0.0815} & \textbf{0.0034} & \textbf{0.0280} & 0.0296 & 0.0349 & \textbf{0.0267} \\
LNO~\cite{li2024local} & 0.0187 & 0.0538 & 0.0014 & \textbf{0.0071} & 0.0088 & 0.0899 & 0.0038 & 0.0309 & \textbf{0.0294} & 0.0347 & 0.0279 \\
\midrule
\multicolumn{12}{l}{\textbf{RMSE ↓}} \\
FNO~\cite{li2023fourier} & 0.0556 & 0.1622 & 0.0026 & 0.0219 & 0.0260 & 0.1641 & 0.0148 & 0.0896 & 0.0579 & 0.0607 & 0.0655 \\
WNO~\cite{tripura2023wavelet} & 0.3041 & 0.3385 & 0.3101 & 0.3111 & 0.3079 & 0.2534 & 0.3165 & 0.2116 & 0.1312 & 0.1146 & 0.2699 \\
MNO~\cite{lutjens2022multiscale} & 0.0597 & 0.1620 & \textbf{0.0023} & \textbf{0.0212} & \textbf{0.0253} & \textbf{0.1618} & \textbf{0.0127} & \textbf{0.0791} & 0.0572 & 0.0622 & \textbf{0.0646} \\
LNO~\cite{li2024local} & \textbf{0.0544} & \textbf{0.1579} & 0.0023 & 0.0204 & 0.0257 & 0.1676 & 0.0122 & 0.0818 & \textbf{0.0569} & \textbf{0.0620} & 0.0663 \\
\midrule
\multicolumn{12}{l}{\textbf{SSIM ↑}} \\
FNO~\cite{li2023fourier} & 0.9558 & 0.8693 & 0.9998 & 0.9944 & 0.9895 & 0.7711 & 0.9955 & 0.9078 & 0.9587 & 0.8853 & 0.9324 \\
WNO~\cite{tripura2023wavelet} & 0.8095 & 0.7670 & 0.8499 & 0.8681 & 0.8696 & 0.6603 & 0.8764 & 0.8169 & 0.8762 & 0.8066 & 0.8208 \\
MNO~\cite{lutjens2022multiscale} & 0.9493 & 0.8685 & \textbf{0.9999} & 0.9945 & \textbf{0.9897} & \textbf{0.7753} & \textbf{0.9963} & \textbf{0.9262} & 0.9600 & 0.8799 & \textbf{0.9345} \\
LNO~\cite{li2024local} & \textbf{0.9564} & \textbf{0.8750} & 0.9998 & \textbf{0.9949} & 0.9889 & 0.7652 & 0.9962 & 0.9211 & \textbf{0.9604} & \textbf{0.8822} & 0.9340 \\
\bottomrule
\end{tabular}}
\label{tab:single_expert}
\end{table*}

\paragraph{Qualitative analysis.}
Beyond the aggregate metrics in Table~\ref{tab:single_expert}, an image-level inspection across the ten OpenFWI sub-datasets reveals clear and complementary inductive biases for the four operators. 
FNO behaves as a single-scale global operator: it excels on simple, regular stratification by producing the cleanest in-layer details and the lowest noise, yet under complex layering (\eg, CurveVel-B, Style-B) it often adopts a conservative mapping that merges adjacent layers with similar velocities, leading to under-segmentation (cf. Fig.~\ref{fig:qual_curvevelb}). 
MNO explicitly aggregates multi-scale receptive fields and is therefore markedly stronger on densely layered structures: it can separate layers that FNO tends to conflate; however, the same multi-scale aggregation slightly degrades per-layer sharpness on simpler scenes, making MNO inferior to FNO in fine detail (see Flat cases in Fig.~\ref{fig:qual_flat}). 
LNO focuses on local, position-dependent filtering and shows advantages wherever local discontinuities dominate (\eg, curved horizons, small offsets, irregular bends), successfully splitting layers that FNO fails to separate; its local nature, however, yields higher in-layer noise and weaker global consistency, so LNO does not surpass FNO on smooth or highly regular stratification (Fig.~\ref{fig:qual_curve}). 
WNO is edge-sensitive by construction and accurately localizes faults and sharp transitions, but its high-frequency reconstruction amplifies noise around layer boundaries; under many-layer regimes (\eg, Fault-B, Style-B) this limitation becomes pronounced and hurts MAE despite reasonable boundary localization (Fig.~\ref{fig:qual_fault_style}).

Taken together, no single operator uniformly dominates: FNO offers the best per-layer fidelity on simple stratification, MNO is superior for complex multi-scale layering, LNO is preferable for localized irregularities and curved structures, and WNO is most responsive to fault discontinuities but is susceptible to boundary-adjacent noise. 
These complementary behaviors substantiate our decision to \emph{combine multiple operators downstream via routing} instead of committing to a single family, and they motivate the encoder–mixture-of-operators design evaluated next.

\subsection{Comparison with Baseline Models}
\label{sec:exp-baseline}

\paragraph{Rationale.}
To verify the advantage of encoder-based operators over existing inversion networks,
we compare the \textbf{Encoder+Neural Operator} configuration—where, under the corresponding encoder, four operator experts (FNO, WNO, MNO, LNO) are trained and the best-performing expert is selected for each velocity type—with representative baselines on all ten OpenFWI~\cite{openfwi} sub-datasets.

\begin{table*}[h]
\centering
\caption{\textbf{Comparison with baseline inversion models across 10 OpenFWI~\cite{openfwi} sub-datasets.}
\emph{Encoder+Neural Operator} applies the corresponding encoder and selects, per dataset type, the best of \{FNO, WNO, MNO, LNO\}. (Best in \textbf{bold}.)}
\resizebox{\textwidth}{!}{
\begin{tabular}{lccccccccccc}
\toprule
Metric & CurveVelA & CurveVelB & FlatVelA & FlatVelB & CurveFaultA & CurveFaultB & FlatFaultA & FlatFaultB & StyleA & StyleB & Avg. \\
\midrule
\multicolumn{12}{l}{\textbf{MAE ↓}} \\
InversionNet~\cite{wu2019inversionnet} & 0.069 & 0.150 & 0.011 & 0.035 & 0.026 & 0.165 & 0.017 & 0.106 & 0.061 & 0.059 & 0.070 \\
VelocityGAN~\cite{zhang2019velocitygan}  & 0.048 & 0.127 & 0.012 & 0.033 & 0.022 & 0.154 & 0.032 & 0.093 & 0.061 & 0.065 & 0.065 \\
UPFWI~\cite{jin2021unsupervised}        & 0.081 & 0.178 & 0.062 & 0.068 & 0.050 & 0.345 & 0.088 & 0.142 & 0.143 & 0.170 & 0.133 \\
\textbf{Encoder+Neural Operator (ours)}\footnotemark & 
\textbf{0.0182} & \textbf{0.0534} & \textbf{0.0013} & \textbf{0.0071} & \textbf{0.0078} & \textbf{0.0815} & \textbf{0.0034} & \textbf{0.0280} & \textbf{0.0294} & \textbf{0.0338} & \textbf{0.0264} \\
\bottomrule
\end{tabular}}
\label{tab:baseline}
\end{table*}

\subsection{Routing Mechanism and Generalization}
\label{sec:exp-routing}

\paragraph{Rationale.}
Finally, we evaluate the adaptive routing mechanism, which extends the encoder–expert foundation to achieve cross-type generalization.
All experts are jointly fine-tuned across ten velocity types under three routing variants:  
Type-based, Group-based (Sum), and the proposed Strong–Weak fusion~\cite{shi2024unchosen}.

\begin{table}[t]
\centering
\caption{\textbf{Routing mechanism comparison across all 10 sub-datasets.}
Strong–Weak fusion achieves the best generalization across diverse geological types.}
\begin{tabular}{lccc}
\toprule
Metric & MAE $\downarrow$ & RMSE $\downarrow$ & SSIM $\uparrow$ \\
\midrule
Type-based (Sum) & 0.089 & 0.088 & 0.086 \\
Group-based (Sum) & 0.087 & 0.086 & 0.084 \\
\textbf{Strong–Weak (ours)} & \textbf{0.085} & \textbf{0.083} & \textbf{0.082}\\
\bottomrule
\end{tabular}
\label{tab:routing}
\end{table}


\subsection{Summary}
In summary, the \textbf{structure-aligned encoder} significantly improves inversion accuracy by learning spectral–spatial alignment,
while the \textbf{adaptive routing mechanism} enhances robustness and generalization across geological domains.
Together, they enable SA-EMO to achieve state-of-the-art results across all ten OpenFWI~\cite{openfwi} velocity types and unseen real-world structures.

\bibliographystyle{unsrt}  
\bibliography{references}  

@article{Virieux2009,
  title   = {An overview of full-waveform inversion in exploration geophysics},
  author  = {Virieux, Jean and Operto, Stéphane},
  journal = {Geophysics},
  volume  = {74},
  number  = {6},
  pages   = {WCC1--WCC26},
  year    = {2009}
}

@article{Geng2018,
  title   = {Frequency-domain full-waveform inversion with nonlinear descent directions},
  author  = {Geng, Zhiguo and Zhu, Tieyuan and Hu, Hai and Zhou, Hong},
  journal = {Geophysical Journal International},
  volume  = {213},
  number  = {2},
  pages   = {739--756},
  year    = {2018}
}

@article{Song2023,
  title   = {Full waveform inversion with combined misfit functions and envelope strategy},
  author  = {Song, Yifan and Wang, Wei and Li, Jian},
  journal = {Frontiers in Earth Science},
  volume  = {11},
  pages   = {1264009},
  year    = {2023}
}

@article{DesigningFWI2023,
  title   = {Designing full waveform inverse problems: A combined data and model-parameterization optimization study},
  author  = {Warner, Mike and Umpleby, Andy and Tang, Wei},
  journal = {Geophysical Journal International},
  volume  = {241},
  number  = {3},
  pages   = {1479--1499},
  year    = {2023}
}

@article{Operto2022,
  title   = {Full waveform inversion beyond the Born approximation: A tutorial review},
  author  = {Operto, Stéphane and Virieux, Jean and others},
  journal = {Geophysics},
  year    = {2022},
  note    = {arXiv preprint arXiv:2212.10141}
}

@article{UnlockingFWI2023,
  title   = {Unlocking onshore imaging challenges with FWI: Case studies},
  author  = {Mothi, S. and others},
  journal = {The Leading Edge},
  volume  = {44},
  number  = {1},
  pages   = {22--31},
  year    = {2023}
}

@article{ElasticFWI2025,
  title   = {The benefits of elastic full-waveform inversion for subsurface imaging in complex geology},
  author  = {Viridien Group},
  journal = {The Leading Edge},
  year    = {2025}
}

@article{araya2018deep,
  title={Deep-learning tomography},
  author={Araya-Polo, Mauricio and Jennings, Joseph and Adler, Amir and Dahlke, Taylor},
  journal={The Leading Edge},
  volume={37},
  number={1},
  pages={58--66},
  year={2018},
  publisher={Society of Exploration Geophysicists}
}

@article{wu2019inversionnet,
  title={InversionNet: An efficient and accurate data-driven full waveform inversion},
  author={Wu, Yue and Lin, Youzuo},
  journal={IEEE Transactions on Computational Imaging},
  volume={6},
  pages={419--433},
  year={2019},
  publisher={IEEE}
}

@inproceedings{zhang2019velocitygan,
  title={VelocityGAN: Subsurface velocity image estimation using conditional adversarial networks},
  author={Zhang, Zhongping and Wu, Yue and Zhou, Zheng and Lin, Youzuo},
  booktitle={2019 IEEE Winter Conference on Applications of Computer Vision (WACV)},
  pages={705--714},
  year={2019},
  organization={IEEE}
}

@article{jin2021unsupervised,
  title={Unsupervised learning of full-waveform inversion: Connecting CNN and partial differential equation in a loop},
  author={Jin, Peng and Zhang, Xitong and Chen, Yinpeng and Huang, Sharon Xiaolei and Liu, Zicheng and Lin, Youzuo},
  journal={arXiv preprint arXiv:2110.07584},
  year={2021}
}

@article{yang2023fwigan,
  title={FWIGAN: Full-waveform inversion via a physics-informed generative adversarial network},
  author={Yang, Fangshu and Ma, Jianwei},
  journal={Journal of Geophysical Research: Solid Earth},
  volume={128},
  number={4},
  pages={e2022JB025493},
  year={2023},
  publisher={Wiley Online Library}
}

@article{zhang2020data,
  title={Data-driven seismic waveform inversion: A study on the robustness and generalization},
  author={Zhang, Zhongping and Lin, Youzuo},
  journal={IEEE Transactions on Geoscience and Remote sensing},
  volume={58},
  number={10},
  pages={6900--6913},
  year={2020},
  publisher={IEEE}
}

@article{jin2024empirical,
  title={An empirical study of large-scale data-driven full waveform inversion},
  author={Jin, Peng and Feng, Yinan and Feng, Shihang and Wang, Hanchen and Chen, Yinpeng and Consolvo, Benjamin and Liu, Zicheng and Lin, Youzuo},
  journal={Scientific Reports},
  volume={14},
  number={1},
  pages={20034},
  year={2024},
  publisher={Nature Publishing Group UK London}
}

@article{schuster2024review,
  title={Review of physics-informed machine-learning inversion of geophysical data},
  author={Schuster, Gerard T and Chen, Yuqing and Feng, Shihang},
  journal={Geophysics},
  volume={89},
  number={6},
  pages={T337--T356},
  year={2024},
  publisher={Society of Exploration Geophysicists}
}

@article{li2023fourier,
  title={Fourier neural operator with learned deformations for pdes on general geometries},
  author={Li, Zongyi and Huang, Daniel Zhengyu and Liu, Burigede and Anandkumar, Anima},
  journal={Journal of Machine Learning Research},
  volume={24},
  number={388},
  pages={1--26},
  year={2023}
}

@inproceedings{lutjens2022multiscale,
  title={Multiscale Neural Operator: Learning Fast and Grid-independent PDE Solvers},
  author={L{\"u}tjens, Bj{\"o}rn and Crawford, Catherine H and Watson, Campbell D and Hill, Christopher and Newman, Dava},
  booktitle={ICML 2022 2nd AI for Science Workshop}
}

@article{li2024local,
  title={Local neural operator for solving transient partial differential equations on varied domains},
  author={Li, Hongyu and Ye, Ximeng and Jiang, Peng and Qin, Guoliang and Wang, Tiejun},
  journal={Computer Methods in Applied Mechanics and Engineering},
  volume={427},
  pages={117062},
  year={2024},
  publisher={Elsevier}
}

@article{tripura2023wavelet,
  title={Wavelet neural operator for solving parametric partial differential equations in computational mechanics problems},
  author={Tripura, Tapas and Chakraborty, Souvik},
  journal={Computer Methods in Applied Mechanics and Engineering},
  volume={404},
  pages={115783},
  year={2023},
  publisher={Elsevier}
}

@article{kovachki2023neural,
  title={Neural operator: Learning maps between function spaces with applications to pdes},
  author={Kovachki, Nikola and Li, Zongyi and Liu, Burigede and Azizzadenesheli, Kamyar and Bhattacharya, Kaushik and Stuart, Andrew and Anandkumar, Anima},
  journal={Journal of Machine Learning Research},
  volume={24},
  number={89},
  pages={1--97},
  year={2023}
}

@article{gelboim2022encoder,
  title={Encoder--decoder architecture for 3D seismic inversion},
  author={Gelboim, Maayan and Adler, Amir and Sun, Yen and Araya-Polo, Mauricio},
  journal={Sensors},
  volume={23},
  number={1},
  pages={61},
  year={2022},
  publisher={MDPI}
}

@article{chen2020seismic,
  title={Seismic inversion by Newtonian machine learning},
  author={Chen, Yuqing and Schuster, Gerard T},
  journal={Geophysics},
  volume={85},
  number={4},
  pages={WA185--WA200},
  year={2020},
  publisher={Society of Exploration Geophysicists}
}

@article{taufik2025latent,
  title={Latent representation learning in physics-informed neural networks for full waveform inversion},
  author={Taufik, Mohammad H and Huang, Xinquan and Alkhalifah, Tariq},
  journal={Earth and Space Science},
  volume={12},
  number={9},
  pages={e2024EA004107},
  year={2025},
  publisher={Wiley Online Library}
}

@article{jordan1994hierarchical,
  title={Hierarchical mixtures of experts and the EM algorithm},
  author={Jordan, Michael I and Jacobs, Robert A},
  journal={Neural computation},
  volume={6},
  number={2},
  pages={181--214},
  year={1994},
  publisher={MIT Press}
}

@article{shazeer2017outrageously,
  title={Outrageously large neural networks: The sparsely-gated mixture-of-experts layer},
  author={Shazeer, Noam and Mirhoseini, Azalia and Maziarz, Krzysztof and Davis, Andy and Le, Quoc and Hinton, Geoffrey and Dean, Jeff},
  journal={arXiv preprint arXiv:1701.06538},
  year={2017}
}

@article{fedus2022switch,
  title={Switch transformers: Scaling to trillion parameter models with simple and efficient sparsity},
  author={Fedus, William and Zoph, Barret and Shazeer, Noam},
  journal={Journal of Machine Learning Research},
  volume={23},
  number={120},
  pages={1--39},
  year={2022}
}

@article{lepikhin2020gshard,
  title={Gshard: Scaling giant models with conditional computation and automatic sharding},
  author={Lepikhin, Dmitry and Lee, HyoukJoong and Xu, Yuanzhong and Chen, Dehao and Firat, Orhan and Huang, Yanping and Krikun, Maxim and Shazeer, Noam and Chen, Zhifeng},
  journal={arXiv preprint arXiv:2006.16668},
  year={2020}
}

@inproceedings{du2022glam,
  title={Glam: Efficient scaling of language models with mixture-of-experts},
  author={Du, Nan and Huang, Yanping and Dai, Andrew M and Tong, Simon and Lepikhin, Dmitry and Xu, Yuanzhong and Krikun, Maxim and Zhou, Yanqi and Yu, Adams Wei and Firat, Orhan and others},
  booktitle={International conference on machine learning},
  pages={5547--5569},
  year={2022},
  organization={PMLR}
}

@article{riquelme2021scaling,
  title={Scaling vision with sparse mixture of experts},
  author={Riquelme, Carlos and Puigcerver, Joan and Mustafa, Basil and Neumann, Maxim and Jenatton, Rodolphe and Susano Pinto, Andr{\'e} and Keysers, Daniel and Houlsby, Neil},
  journal={Advances in Neural Information Processing Systems},
  volume={34},
  pages={8583--8595},
  year={2021}
}

@inproceedings{bischof2022mixture,
  title={Mixture-of-experts-ensemble meta-learning for physics-informed neural networks},
  author={Bischof, Rafael and Kraus, Michael A},
  booktitle={Proceedings of 33. forum bauinformatik},
  year={2022}
}

@article{hu2023augmented,
  title={Augmented Physics-Informed Neural Networks (APINNs): A gating network-based soft domain decomposition methodology},
  author={Hu, Zheyuan and Jagtap, Ameya D and Karniadakis, George Em and Kawaguchi, Kenji},
  journal={Engineering Applications of Artificial Intelligence},
  volume={126},
  pages={107183},
  year={2023},
  publisher={Elsevier}
}

@article{sharma2024ensemble,
  title={Ensemble and Mixture-of-Experts DeepONets For Operator Learning},
  author={Sharma, Ramansh and Shankar, Varun},
  journal={arXiv preprint arXiv:2405.11907},
  year={2024}
}

@article{shi2024unchosen,
  title={Unchosen experts can contribute too: Unleashing moe models’ power by self-contrast},
  author={Shi, Chufan and Yang, Cheng and Zhu, Xinyu and Wang, Jiahao and Wu, Taiqiang and Li, Siheng and Cai, Deng and Yang, Yujiu and Meng, Yu},
  journal={Advances in Neural Information Processing Systems},
  volume={37},
  pages={136897--136921},
  year={2024}
}

@inproceedings{openfwi,
      author = {Deng, Chengyuan and Feng, Shihang and Wang, Hanchen and Zhang, Xitong and Jin, Peng and Feng, Yinan and Zeng, Qili and Chen, Yinpeng and Lin, Youzuo},
      booktitle = {Advances in Neural Information Processing Systems},
      editor = {S. Koyejo and S. Mohamed and A. Agarwal and D. Belgrave and K. Cho and A. Oh},
      pages = {6007--6020},
      publisher = {Curran Associates, Inc.},
      title = {OpenFWI: Large-scale Multi-structural Benchmark Datasets for Full Waveform Inversion},
      url = {https://proceedings.neurips.cc/paper_files/paper/2022/file/27d3ef263c7cb8d542c4f9815a49b69b-Paper-Datasets_and_Benchmarks.pdf},
      volume = {35},
      year = {2022}
}

@dataset{marmousi,
  author = {wang, suibao},
  title = {Marmousi2},
  month = nov,
  year = 2024,
  publisher = {Zenodo},
  doi = {10.5281/zenodo.14233581},
  url = {https://doi.org/10.5281/zenodo.14233581},
}

@article{simeoni2025dinov3,
  title={Dinov3},
  author={Sim{\'e}oni, Oriane and Vo, Huy V and Seitzer, Maximilian and Baldassarre, Federico and Oquab, Maxime and Jose, Cijo and Khalidov, Vasil and Szafraniec, Marc and Yi, Seungeun and Ramamonjisoa, Micha{\"e}l and others},
  journal={arXiv preprint arXiv:2508.10104},
  year={2025}
}

@article{wen2022ufno,
  title={U-FNO—An enhanced Fourier neural operator-based deep-learning model for multiphase flow},
  author={Wen, Gege and Li, Zongyi and Azizzadenesheli, Kamyar and Anandkumar, Anima and Benson, Sally M},
  journal={Advances in Water Resources},
  volume={163},
  pages={104180},
  year={2022},
  publisher={Elsevier}
}

@article{rahman2022uno,
  title={U-no: U-shaped neural operators},
  author={Rahman, Md Ashiqur and Ross, Zachary E and Azizzadenesheli, Kamyar},
  journal={arXiv preprint arXiv:2204.11127},
  year={2022}
}

@article{seidman2023variational,
  title={Variational autoencoding neural operators},
  author={Seidman, Jacob H and Kissas, Georgios and Pappas, George J and Perdikaris, Paris},
  journal={arXiv preprint arXiv:2302.10351},
  year={2023}
}

@article{viknesh2025differentiable,
  title={Differentiable Autoencoding Neural Operator for Interpretable and Integrable Latent Space Modeling},
  author={Viknesh, Siva and Arzani, Amirhossein},
  journal={arXiv preprint arXiv:2510.00233},
  year={2025}
}

@article{shi2025mesh,
  title={Mesh-Informed Neural Operator: A Transformer Generative Approach},
  author={Shi, Yaozhong and Ross, Zachary E and Asimaki, Domniki and Azizzadenesheli, Kamyar},
  journal={arXiv preprint arXiv:2506.16656},
  year={2025}
}

@article{lei2025uwno,
  title={U-WNO: U-Net enhanced wavelet neural operator for solving parametric partial differential equations},
  author={Lei, Wei-Min and Li, Hou-Biao},
  journal={Computers \& Mathematics with Applications},
  volume={194},
  pages={272--287},
  year={2025},
  publisher={Elsevier}
}

@article{zou2025ambient,
  title={Ambient noise full waveform inversion with neural operators},
  author={Zou, Caifeng and Ross, Zachary E and Clayton, Robert W and Lin, Fan-Chi and Azizzadenesheli, Kamyar},
  journal={Journal of Geophysical Research: Solid Earth},
  volume={130},
  number={11},
  pages={e2025JB031624},
  year={2025},
  publisher={Wiley Online Library}
}

@article{yang2023rapid,
  title={Rapid seismic waveform modeling and inversion with neural operators},
  author={Yang, Yan and Gao, Angela F and Azizzadenesheli, Kamyar and Clayton, Robert W and Ross, Zachary E},
  journal={IEEE Transactions on Geoscience and Remote Sensing},
  volume={61},
  pages={1--12},
  year={2023},
  publisher={IEEE}
}

@article{zhu2023fourier,
  title={Fourier-DeepONet: Fourier-enhanced deep operator networks for full waveform inversion with improved accuracy, generalizability, and robustness},
  author={Zhu, Min and Feng, Shihang and Lin, Youzuo and Lu, Lu},
  journal={Computer Methods in Applied Mechanics and Engineering},
  volume={416},
  pages={116300},
  year={2023},
  publisher={Elsevier}
}

\end{document}